\documentclass[letterpaper, 10pt, journal]{ieeeconf}  

\IEEEoverridecommandlockouts                              

\usepackage{times}
\usepackage{dsfont}
\usepackage{amsmath}
\usepackage{amssymb}
\usepackage{multirow}
\usepackage{mathtools}
\usepackage{units}
\usepackage{float}
\usepackage[]{units}
\usepackage{xcolor}
\usepackage{url}
\usepackage{dsfont}
\usepackage{algorithm}
\usepackage{algpseudocode}
\usepackage{soul} 
\usepackage[makeroom]{cancel}
\usepackage[printonlyused]{acronym}
\usepackage{hyperref}
\usepackage{tikz}
\usepackage{pgfplots}

\usepackage{multicol}
\usepackage{graphicx}  
\usepackage{tabularx}  
\usepackage{colortbl}  
\usepackage{booktabs}  
\usepackage{subcaption}

\newacro{AOW}{ANYmal On Wheels}
\newacro{MDP}{Markov Decision Process}
\newacro{CMDP}{Constrained Markov Decision Process}
\newacro{RL}{Reinforcement Learning}
\newacro{PPO}{Proximal Policy Optimization}
\newacro{CPPO}{Constrained Proximal Policy Optimization}

\newacro{MPC}{Model Predictive Control}
\newacro{COT}{Cost of Transport}
\newacro{VAE}{Variational Autoencoder}
\newacro{IMU}{Inertial Measurement Unit}
\newacro{SEA}{Series Elastic Actuator}
\newacro{MLP}{Multi Layer Perceptron}
\newacro{RNN}{Recurrent Neural Network}
\newacro{GRU}{Gated Recurrent Unit}
\newacro{CNN}{Convolutional Neural Network}
\newacro{SAC}{Soft Actor Critic}

\newacro{ReLU}{Rectified Linear Unit}
\newacro{GAE}{Generalized Advantage Estimation}
\newacro{P3O}{Penalized Proximal Policy Optimization}
\newacro{IPO}{Interior-point Policy Optimization}
\newacro{CRPO}{Constraint-Rectified Policy Optimization}
\newacro{FOCOPS}{First-Order Constrained Optimization in Policy Space}
\newacro{NP3O}{Normalized Penalty Proximal Policy Optimization}
\newacro{TRPO}{Trust Region Policy Optimization}
\newacro{CPO}{Constrained Policy Optimization}
\newacro{PCPO}{Projection-Based Constrained Policy Optimization}

\title{\LARGE \bf
Evaluation of Constrained Reinforcement Learning Algorithms \\ for Legged Locomotion
}

\author{Joonho Lee$^{*}$, Lukas Schroth$^{*}$, Victor Klemm, Marko Bjelonic, Alexander Reske, and Marco Hutter
\thanks{* \textit{Joonho Lee and Lukas Schroth contributed equally.}
All authors are with the Robotic Systems Lab (RSL), ETH Zürich, Switzerland. {\tt\footnotesize jolee@ethz.ch}
}
\thanks{This work has been submitted to the IEEE for possible publication. Copyright may be transferred without notice, after which this version may no longer be accessible.}
}

\begin{document}

\maketitle
\thispagestyle{empty}
\pagestyle{empty}

\begin{abstract}
Shifting from traditional control strategies to Deep Reinforcement Learning (RL) for legged robots poses inherent challenges, especially when addressing real-world physical constraints during training. While high-fidelity simulations provide significant benefits, they often bypass these essential physical limitations. In this paper, we experiment with the Constrained Markov Decision Process (CMDP) framework instead of the conventional unconstrained RL for robotic applications. We perform a comparative study of different constrained policy optimization algorithms to identify suitable methods for practical implementation. Our robot experiments demonstrate the critical role of incorporating physical constraints, yielding successful sim-to-real transfers, and reducing operational errors on physical systems. The CMDP formulation streamlines the training process by separately handling constraints from rewards. Our findings underscore the potential of constrained RL for the effective development and deployment of learned controllers in robotics.
\end{abstract}

\section{INTRODUCTION}

The use of Deep \ac{RL} for robotic control is on the rise, revolutionizing the way control policies are created for legged robots and other complex dynamic systems. 
Particularly, model-free approaches have gained prominence, replacing traditional optimization-based methods. 
This paradigm shift can be attributed to the high-capacity neural network models, effective model-free algorithms that can solve complex problems, and efficient tools for data-generation (i.e. simulations). 
As a result, the synthesis of locomotion policies for legged robots has become more straightforward and accessible, as evidenced by the growing number of \ac{RL}-based controllers in recent literature.

The so-called sim-to-real approach is commonly employed, where policy training solely relies on simulated data.
This is due to the inherent requirements of widely-used algorithms such as \ac{PPO}~\cite{schulman2017proximal} and \ac{SAC}~\cite{haarnoja2018soft}, which demand random exploration and a significant number of samples. As a result, training policies directly on hardware is both impractical and hazardous.
In recent years, diverse approaches have emerged to enhance simulation fidelity (e.g., actuator modeling~\cite{tan2018sim}, hybrid simulator~\cite{hwangbo2019learning, jiang2021simgan}), and to robustify policies against domain shifts (e.g., dynamics randomization~\cite{peng2018sim,xie2021dynamics}, privileged training~\cite{lee2020learning}).

\begin{figure}
    \centering
    \includegraphics[width=\linewidth]{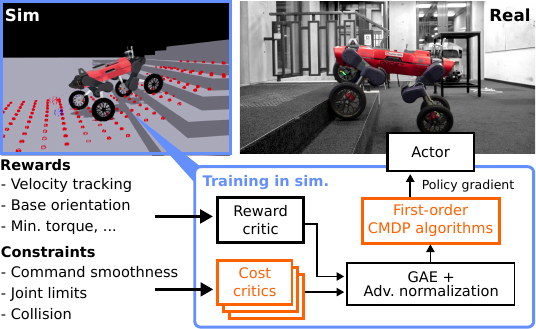}
        \caption{Wheeled-legged locomotion trained via constrained policy optimization. Additional components to conventional PPO are highlighted.}
    \label{fig1}
\end{figure} 

Notably, while most existing research emphasizes enhancing simulation accuracy and regularizing policies for sim-to-real transfer, a gap persists in the literature — a lack of attention to physical constraints. Despite the studies done in understanding and simulating the physical properties of hardware, the incorporation of essential physical constraints during training remains under-explored. 

These constraints can be physical, such as limits on joint velocities, torque limits, or safety regulations.
Considering such constraints is a common practice in model-based approaches~\cite{grandia2019frequency,kang2022nonlinear}.
Existing literature provides compelling evidence of its significance. For instance, Gangapurwala et al.~\cite{gangapurwala2020guided} first utilized a \ac{CPPO} algorithm to train a locomotion controller for a quadrupedal robot, achieving both constraint-consistency and high performance.
Kim et al.~\cite{kim2023not} also experimented with a modified version of \ac{IPO}~\cite{liu2020ipo} algorithm and showed rough-terrain locomotion with a generalizable \ac{CMDP} formulation.

In this paper, we evaluate various first-order constrained policy optimization methods, focused on the application to legged locomotion.
We formulate velocity-tracking locomotion as a \ac{CMDP}~\cite{altman2021constrained}, effectively isolating the physical constraints from the reward function.
Additionally, we introduce a modification to existing algorithms to enhance both stability and final performance. 
 
Our main results can be summarized as follows:
\begin{enumerate}
    \item We conduct a comprehensive comparison of first-order constrained RL algorithms and select the most suitable one for practical applications based on constraint violations and final performance.

    \item We demonstrate the effectiveness of the constrained \ac{RL} approach in handling physical constraints with the wheeled-legged robot shown in Fig.~\ref{fig1}.
\end{enumerate}

From our experiments, we found out that the constrained \ac{RL} formulation yields fewer constraint violations compared to the commonly used unconstrained approach. Additionally, this reduces the reward-shaping effort for physical limitations, a common practice in the existing research.

* This is a preprint. We will publish our implementations of the algorithm in \url{https://github.com/junja94/cmdp_ppos} with the final version of the paper.

\section{BACKGROUND}

\subsection{Constrained Policy Optimization} \label{def_cmdp}
In \ac{RL}, a control problem is typically modeled as a \ac{MDP}, which is described by a tuple $(S, A, r, p, \mu)$. 
Here, $S$ is the set of states, $A$ is the set of Actions, $r: S \times A \times S \rightarrow \mathbb{R}$ is the reward function, $p: S \times A \times S \rightarrow [0,1]$ is the state transition probability and $\mu$ is the initial state distribution. To solve an \ac{MDP}, we aim to find a policy $\pi: S \mapsto \mathcal{P}(A)$ that maximizes
\begin{equation}
	\label{eq_return}
	J_R(\pi) = {\mathbb{E}}\left[\sum_{t=0}^{\infty} \gamma^t r(s_t,a_t,s_{t+1})\right],
\end{equation}
where $\gamma \in [0, 1)$ is the discount factor.
Here, the expectation $\mathbb{E}[\ldots]$ represents the empirical average over a finite batch of samples. $s_0$ is sampled from an initial state distribution $\mu$ and trajectories sampled using $\pi$. 

To address constrained problems, this framework is extended into a \ac{CMDP}. The \ac{MDP} is augmented with a set $C$ of cost functions that capture constraint violations $\{c_1, c_2, \ldots, c_n\}$ and  corresponding limits $E = \{\epsilon_1, \epsilon_2, \ldots, \epsilon_n\}$~\cite{altman2021constrained, achiam2017constrained}.
Each $c_i: S \times A \times S \rightarrow \mathbb{R}$ maps state-action-state triplets to the cost of the state transition.
In the constrained setting, an optimal policy maximizes the expected discounted return in Eq.~\ref{eq_return}, while keeping the discounted sum of future costs $c_i$ below their respective threshold $\epsilon_i$, yielding the constrained optimization problem:
\begin{equation}
	\label{eq_objectiveCMDP}
	\begin{aligned}
		\max_{\pi} \quad & J_R(\pi) \\
		\text{s.t.} \quad & \forall i \in \{1,\ldots,n\}, \ J_{C_i}(\pi) \leq \epsilon_i,
	\end{aligned}
\end{equation}
where
\begin{equation}
	J_{C_i}(\pi) = \mathbb{E}\left[\sum_{t=0}^{\infty} \gamma^t c_i(s_t,a_t,s_{t+1})\right].
\end{equation}
While many constrained \ac{RL} problems in the literature consider a single constraint (e.g \cite{shen2022penalized, achiam2017constrained, ray2019benchmarking}), the \ac{CMDP} framework is not limited to the single constraint setup~\cite{kim2023not}.

Derived by the performance difference lemma by Shen et al.~\cite{shen2022penalized}, the constrained optimization problem in Eq.~\ref{eq_objectiveCMDP} can be reformulated as follows:
\begin{subequations}
	\label{eq_equivOptimCMDP}
	\begin{align}
		\max_{\pi'} \quad & 
  {\mathbb{E}}\bigg[A_{R,t}^\pi(s,a)\bigg] \\
		\label{eq_reformConstr}
		\text{s.t.} \quad & \underbrace{J_{C_i}(\pi) + \frac{1}{1-\gamma}
  {\mathbb{E}_{\pi'}}\bigg[A_{{C_i},t}^\pi(s,a)\bigg]}_{J_{C_i}(\pi')} \leq \epsilon_i \quad \forall i.
	\end{align}
\end{subequations}
where $A_{R,t}^\pi(s,a)$ and  $A_{{C_i},t}^\pi(s,a)$ are estimators of the reward advantage function and cost advantage function for the $i$-th constraint at timestep $t$, respectively.


\subsection{First-order Optimization Methods for CMDPs}
We compare five first-order policy optimization algorithms in order to identify a method that is performant and stable.
As higher-order algorithms typically require
resource-intensive computation of the inverse Hessian or inverse Hessian-vector products (see, e.g., CPO~\cite{achiam2017constrained}, PCPO~\cite{yang2020projection}, TRPO-Lagrangian~\cite{ray2019benchmarking}), we restrict our scope to first-order algorithms.
We considered practical aspects such as the number of hyperparameters, availability of an implementation and the presented empirical results.

\subsubsection{{P3O}}
Shen et al.~\cite{shen2022penalized} proposed to augment the \ac{PPO} objective with penalties on the constraint violations.
The objective function for \ac{P3O} is defined as: 
\begin{equation}
 L_R^{\mathrm{CLIP}}(\theta') - \sum_i \kappa_i \cdot \max \left\{0, J_{C_i}(\pi') - \epsilon_{i}\right\},
\end{equation}
$\kappa_i$ controls the weight of each constraint.

The first term $L_R^{\mathrm{CLIP}}(\theta')$ is the clipped surrogate objective by Schulman et al.~\cite{schulman2017proximal}, defined as:
\begin{equation}
    L_R^{\mathrm{CLIP}}(\theta') = 
  {\mathbb{E}} \left[
		\min ( r_t(\theta') \tilde{A}_{R,t}^{\pi_{\theta}}, \operatorname{clip}(r_t(\theta')) \tilde{A}_{R,t}^{\pi_{\theta}}
		\right],
\end{equation}
where $r_t(\theta')$ denote the probability ratio $\frac{\pi'(a_t|s_t)}{\pi(a_t|s_t)}$, and the operation $\operatorname{clip}(\cdot)$ clips the value between $1-\delta$ and $1+\delta$ with $\delta$ controlling the magnitude of policy updates. 
$\tilde{A}_{R,t}$ denotes the normalized reward advantage.

Similarly, the final objective of P3O is obtained using importance sampling and clipping of the importance ratios of the cost advantages:
\begin{equation}\label{p3o_obj}
	L^{P3O}(\theta') = L_R^{\mathrm{CLIP}}(\theta') - \sum_i \kappa_i \cdot \max \left\{0, L_{C_i}^{\mathrm{VIOL}}(\theta')\right\},
\end{equation}
with
\begin{subequations}
\begin{equation*}
    L_{C_i}^{\text{VIOL}}(\theta') = L_{C_i}^{\text{CLIP}}(\theta') + (1-\gamma)(J_{C_i}(\pi_{\theta}) - \epsilon_i)
\end{equation*}
\begin{equation*}
    L_{C_i}^{\text{CLIP}}(\theta') =
    {\mathbb{E}} \left[
		\max ( r_t(\theta') {A}_{{C_i},t}^{\pi_{\theta}}, \operatorname{clip}(r_t(\theta')) {A}_{{C_i},t}^{\pi_{\theta}})
		\right].
\end{equation*}

\end{subequations}

\subsubsection{PPO-Lagrangian}
Chow et al.~\cite{chow2017pdo} proposed to utilize the Lagrangian relaxation.
The Lagrangian method approaches constraint problems with objective $f(\theta)$ and constraint $g(\theta)$ by minimizing the Lagrange dual with dual variable $\lambda$, resulting in the unconstrained objective:
	\begin{equation}
		\min_{\lambda \geq 0} \max_{\theta} \mathcal{L}(\theta, \lambda) \doteq f(\theta) - \lambda g(\theta).
	\end{equation}
 
 Approximate solutions of this minimax objective can be obtained via the iterative primal-dual method, which alternates between updates on the primal variable $\theta$ and the dual variable $\lambda$ \cite{liang2018accelerated}. 
 In practice, the updates are typically realized with gradient ascent and descent steps on $\theta$ and $\lambda$, where the other variable is kept fixed. Intuitively, $\lambda$ behaves like a penalty parameter that increases when the constraint is violated and decreases when it is satisfied.
 
OpenAI researchers~\cite{ray2019benchmarking} suggested utilizing the iterative primal-dual method with the PPO objective to derive the following update:
\begin{align}
    \theta' &= \theta + \alpha_{\theta} \nabla_\theta \left(L_R^{\mathrm{CLIP}}(\theta) - \sum_{i}\lambda_i L_{C_i}^{\text {CLIP}}(\theta)\right), \\
    \lambda_i' &= \lambda_i + \alpha_{\lambda_i}(J_{C_i}(\theta) - \epsilon_i). \label{eq:lambda_update_ppolagr}
\end{align}

Here, $\alpha_{\theta}$ and $\alpha_{\lambda_i}$ are the learning rates of the gradient ascent and descent steps, respectively. $\lambda_i'$ is typically cut off at zero, to ensure non-negativity of the penalty parameter.

\subsubsection{IPO}
Inspired by the interior-point method for constrained optimization problems, \ac{IPO} uses logarithm barrier functions $\phi(x) = {\log(-x)}/{k}$, with the hyperparameter $k$ to achieve an infinitely large penalty as the estimated cost returns approach the constraint threshold $\epsilon_i$. This results in the objective:
\begin{equation}\label{eq_ipo}
	L_R^{\mathrm{CLIP}}(\theta') + \sum_{i} \phi(J_{C_i}(\theta') - \epsilon_i),
\end{equation}

where $J_{C_i}(\theta')$ can be estimated based on the advantages using Eq.~\ref{eq_reformConstr}.
\subsubsection{CRPO}
\ac{CRPO}~\cite{xu2021crpo} alternates between maximizing the objective and minimizing the constraint violations whenever the constraints are violated:
\begin{equation}
	L^{CRPO}(\theta') = \mathds{1}_{J_C(\theta) \leq \epsilon_i} \cdot L_R^{\mathrm{CLIP}}(\theta') - \mathds{1}_{J_C(\theta) > \epsilon_i} \cdot L_C^{\mathrm{CLIP}}(\theta').
\end{equation}
\subsubsection{FOCOPS}
\ac{FOCOPS} solves the constrained optimization problem in policy space and then projects the solution back into parameter space, effectively also leading to an objective function with a constraint penalty \cite{zhang2020first}. For a detailed derivation we refer to the original paper of Zhang at al. \cite{zhang2020first}. 

The algorithms \ac{P3O}~\cite{shen2022penalized}, PPO-Lagrangian~\cite{ray2019benchmarking}, and \ac{IPO}~\cite{liu2020ipo} relax the constrained optimization problem in Eq.\ref{eq_objectiveCMDP} into an unconstrained one using additional penalties to the \ac{PPO} objective. 
\ac{CRPO} takes a simpler approach and alternates between \ac{PPO} updates with reward and cost advantages \cite{xu2021crpo}.
FOCOPS~\cite{zhang2020first} solves the constrained optimization problem in policy space.

\section{METHOD}
We define a \ac{CMDP} to train policies for velocity-tracking perceptive locomotion.
The training environment and \ac{MDP} inherit from the quadruped environment by Rudin et al.~\cite{rudin2022learning}.

\subsection{CMDP for Perceptive Locomotion}\label{sec:cmdp}

\subsubsection{Reward Functions}
Our reward function is a sum of different reward terms provided in Table~\ref{tab:rewards}.
We define three categories:

\begin{itemize}
    \item Task Reward: This defines the main task objective. In our experiment, the main task is to track linear velocity command in horizontal direction ($v_{xy}$) and yaw rate ($\omega_z$).
    \item Style Reward: There can be many solutions for the velocity tracking, e.g., different gait, base height, or different orientation.
    We use extra rewards to guide natural-looking motions.
    Kim et al.~\cite{kim2023not} similarly achieved this by applying constraints to gait and other physical quantities.

    \item Constraint Reward:
    High penalty is given when the physical limits are violated. The constraint rewards are replaced by the constraints in \ac{CMDP}.
    
\end{itemize}

\begin{table}
\centering
\begin{tabular}{l|l}

\hline

\multicolumn{2}{c}{Task Rewards} \\ 
\hline
Linear Velocity & $\exp(-2.0 \cdot \lvert\lvert {v^{targ}_{xy}} - v_{xy}\rvert\rvert^2$)\\
Yaw Rate & $ \exp(-2.0 \cdot \lvert\lvert {\omega^{targ}_{z}} - \omega_{z}\rvert\rvert^2$)\\
\hline
\multicolumn{2}{c}{Style Rewards} \\ 
\hline
Base Stability & $ \exp(- v_{z}^2) + \exp(- \lvert\lvert\omega_{x,y}\rvert\rvert^2) $\\
Height & $ -0.5 \ \lvert h^{targ} - h_{robot}\rvert $, $h^{targ} = 0.5$  \\
Joint Torque Minimization &  $1\mathrm{e}{-6} \ \lvert\lvert \tau \rvert \rvert ^2 $ \\

Joint Motion &  $1\mathrm{e}{-5} \ \lvert\lvert \dot{q} \rvert \rvert ^2 + 1\mathrm{e}{-6} \ \lvert\lvert \ddot{q} \rvert \rvert ^2 $ \\
\hline
    \multicolumn{2}{c}{Constraint Rewards (\textbf{Removed for CMDP}) } \\ 
\hline
Command Smoothness 1 &  $- 0.01 \ \lvert\lvert q_{t}^{des} - q_{t-1}^{des} \rvert\rvert^2$ \\
Command Smoothness 2 &  $- 0.01 \ \lvert\lvert q_{t}^{des} - 2q_{t-1}^{des} + q_{t-2}^{des} \rvert\rvert^2$ \\
Joint Torque Limits & $-0.01 \ \sum\max( \lvert \tau_{i,t} \rvert - \tau^{lim}_i, 0) ^2 $ \\
Joint Speed Limits & $-0.1 \ \sum\max( \lvert \dot{q}_{i,t} \rvert - \dot{q}^{lim}_i, 0) ^2 $ \\
Joint Position Upper Limits &  $-10.0 \ \sum \max(q_{i,t} - q^{ub}_i,0)^2$  \\
Joint Position Lower Limits &  $-10.0 \ \sum \max(q^{lb}_i - q_{i,t},0)^2$  \\
Body Contact & $- (\text{Number of non-wheel contacts} )$ \\
\hline
\multicolumn{2}{c}{}\\ 
\end{tabular}
\caption{Reward Functions. $q$ and $\tau$ are joint position and torque vectors. $g^b$ denotes the gravity vector in base frame.}
\label{tab:rewards}
\end{table}

\subsubsection{Constraints}
For all constraints, we set $\epsilon_i = 0$ and defined cost functions such that each cost encapsulates a specific physical quantity:

\begin{itemize}
    \item Command Smoothness:  
    For the sim-to-real transfer, it is crucial to consider the tracking bandwidth of the physical actuators~\cite{grandia2019frequency}.
    Existing works regularize the output with negative rewards on the first or second order derivative of the commands~\cite{lee2020learning,rudin2022learning,kim2023not}. 
    This prevents infeasible commands, reduces sim-to-real discrepancy in the joint space, and vibration on the hardware.
    

    \quad We define two constraint functions as:
    \begin{align*}
        c_{c1,i} &=  \max(0,  \lvert (q_{t,i}^{des} - q_{t-1,i}^{des}) / dt\rvert - \dot{q}^{des,*}) \\
        c_{c2,i} &=\max(0,  \lvert (q_{t,i}^{des} - 2q_{t-1,i}^{des}+q_{t-2,i}^{des}) / dt^2\rvert - \ddot{q}^{des, *})  
    \end{align*}
    for each joints ($i \in {\text{joints}})$. $dt$ is the timestep and $\dot{q}^{des,*}$ and $\ddot{q}^{des,*}$ are thresholds.
    
    \quad The discounted sum of both costs are restricted to be below the desired thresholds by setting  $\epsilon=0.0$. $\dot{q}^{des,*}$ and $\ddot{q}^{des,*}$ are hyperparameters, with $\dot{q}^{des,*}$ set as half of the joint speed limit, and $\ddot{q}^{des,*} = \dot{q}^{des,*} / dt$.


    \item Joint Speed:
    The constraint function is defined as an indicator function:
    \begin{equation*}
    c_{qv} = \mathds{1} ( \sum_{i\in\text{joints}}\mathds{1}({\lvert \dot{q}_{t,i} \rvert  > \dot{q}^{*}_i }) > 0.0 ).
    \end{equation*}
    In other words, $c_{qv} = 1$ if any of the joints violates the speed limitation. $\dot{q}^{*}$ is the physical limit of the actuator.
    \item Joint Torque:
    Joint torque constraint is defined similarly to the joint speed constraint. 
    \begin{equation*}
    c_{\tau} = \mathds{1} ( \sum_{i\in\text{joints}}\mathds{1}({\lvert {\tau_{t,i}} \rvert  > \tau^{*}_i }) > 0.0 ).
    \end{equation*}
    
    \item Joint Position:
    Each joint has different upper bound (${q}^{ub}$) and lower bound (${q}^{lb}$) positions. We only set the limit angle for the hip joints to avoid self-collision.  
    \begin{equation*}
    c_{q} = \mathds{1}( 
    \sum_{i\in\text{hip joints}}(\mathds{1}({q_{t,i} > {q}^{ub}_i})  +
    \mathds{1} ({q_{t,i} < {q}^{lb}_i}) ) 
    > 0.0 ).
    \end{equation*}

    \item Undesirable Body Contact:
    The cost is $1.0$ when there is any contact at the body parts except for the wheel or foot, including self-collision. 
    
\end{itemize}

\subsection{Normalizing Cost Advantages}

Advantage normalization is a widely used heuristic to improve the stability of policy gradient algorithms~\cite{andrychowicz2020matters}.
This technique is also applicable for constrained \ac{RL} algorithms.

Consider the simplified objective for P3O: 
\begin{equation*}
    	L(\theta') = {\mathbb{E}} \left[ r(\theta')\left(A_R^{\theta} - \kappa \cdot A_C^{\theta}\right)\right]
\end{equation*}
The un-normalized advantages $A_R^{\theta}$ and $A_C^{\theta}$ can have different magnitudes, depending on the reward, constraints, and the current policy's behavior.
With normalized advantages, 
\begin{equation}
	L(\theta') = {\mathbb{E}} \left[ r(\theta') \left(\tilde{A}_R^{\theta} - \kappa \cdot \tilde{A}_C^{\theta}\right) \right]
\end{equation}
then the weighting of the constraints ($\kappa$) remains unchanged regardless of the reward and cost functions.
E.g., $\kappa=1$ always corresponds to equal weighting of the reward and cost advantages.
This makes the algorithm more stable and improves generalization across tasks, also as evidenced by Kim et al.~\cite{kim2023not}. 
Furthermore, this prevents the cost advantages from vanishing when cost violation is low.

For \ac{P3O} and \ac{IPO}, we need to reformulate the objectives in Eq.~\ref{p3o_obj} and Eq.~\ref{eq_ipo}.
We start by expressing the constraint in Eq.~\ref{eq_reformConstr} in terms of normalized advantages:
\begin{equation}
    \label{eq_ineq_normalized}
	\frac{(1 -\gamma)(J_{C_i}(\pi) - \epsilon_i) + \mu_{C_i}}{\sigma_{C_i}}
 + {\mathbb{E}}\bigg[
 \underbrace{\frac{A_{{C_i},t}^\pi - \mu_{C_i}}{\sigma_{C_i}}}_{\mathclap{\tilde{A}^{\pi}_{C_i,t}}}\bigg] \leq 0 \quad \forall_i.
\end{equation}
Here, $\mu_{C_i}$, $\sigma_{C_i}$  are the mean and standard deviations of the cost advantages. $\tilde{A}^{C_i}_{\pi}$ denotes the normalized advantages. 
Using importance sampling with clipping, one obtains 
\begin{equation}
    \label{eq_constr_final}
    \resizebox{\linewidth}{!}{
        $L_{C_i}^{\text{VIOL,N}}(\theta') = L_{C_i}^{\text{CLIP,N}}(\theta') + \frac{(1-\gamma)(J_{C_i}(\pi_{\theta}) - \epsilon_i) + \mu_{C_i}}{\sigma_{C_i}} \leq 0.$
    }
\end{equation}
The superscript $N$ indicates the usage of normalized advantage estimates. Penalizing violations of Eq. \ref{eq_constr_final}, leads to the objectives
\begin{align*}
    L^{\text{N-P3O}}(\theta') & = L_R^{\text{CLIP,N}}(\theta') - \sum_i \kappa_i \cdot \max \left\{0, L_{C_i}^{\text{VIOL,N}}(\theta')\right\}, \\
    L^{\text{N-IPO}}(\theta') & = L_R^{\text{CLIP,N}}(\theta') + \sum_{i} \phi(L_{C_i}^{\text{VIOL,N}}(\theta')).
\end{align*}
We will refer to these modified versions of P3O and IPO as N-P3O and N-IPO throughout the rest of the paper.


\begin{figure}
    \centering
        \includegraphics[width=0.95\linewidth]{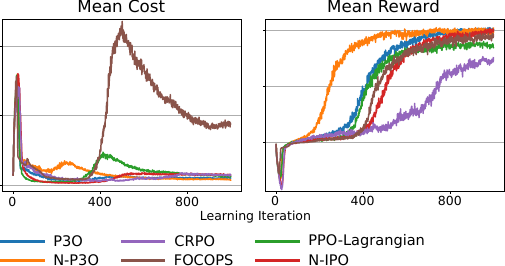}
    \caption{Learning curves of the selected CMDP algorithms.}
    \label{fig:cost_rwd}
\end{figure}

\begin{table}
    \centering
        \begin{tabular}{c | cc}
            \toprule
            & {Reward} & Violations per episode \\
            \midrule
            {PPO (unconstrained)} & 24.96 ($\pm$ 0.67) & 533.44 ($\pm$ 108.94) \\
            \midrule
            {P3O} & 24.13 ($\pm$ 1.55) & 0.96 ($\pm$ 1.35)\\
            {N-P3O} & 24.13 ($\pm$ 1.14) & \textbf{0.49} ($\pm$ 0.88)  \\
            {PPO-Lagrangian}  & 23.68 ($\pm$ 1.87) & 0.99 ($\pm$ 1.31) \\
            {N-IPO} & \textbf{24.67} ($\pm$ 0.84) & 1.33 ($\pm$ 1.69)  \\
            {CRPO}  & 22.28 ($\pm$ 1.70) & 0.96 ($\pm$ 1.22)  \\
            {FOCOPS} & 22.65 ($\pm$ 3.02) & 15.82 ($\pm$ 11.67)  \\
            \bottomrule
        \end{tabular}
    \caption{Final performances of the CMDP algorithms.}
    \label{tab_selected_cmdp_algos}
\end{table}

\begin{figure*}
    \centering
    \includegraphics[width=\linewidth]{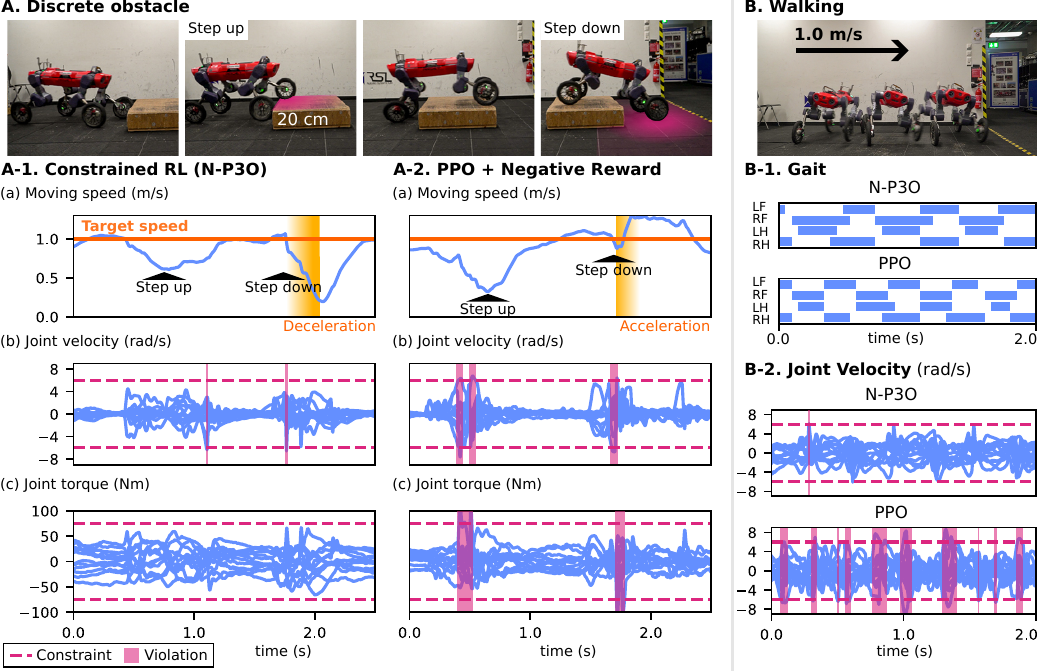}
        \caption{Robot experiments with constraints. (A) Traversing a \unit[20]{cm} high block with \unit[1.0]{m/s} command to the front. (B) Walking in $y$-direction at the maximum speed.}
    \label{fig_robot}
\end{figure*} 

\begin{table*}[tpb]
\centering
\begin{tabular}{c| cccccc}
\toprule
& {Tuning Iteration} & {Parameters} & {Episode reward} & $\#_{\text{violations}}$ / episode  \\
\midrule
{PPO (no constraint)}  & - & - & 24.96 ($\pm$ 0.67) & 533.44 ($\pm$ 108.94) \\
\midrule
\multirow{5}{*}{{P3O}} & 1 & $\kappa=1$ & 25.23 ($\pm$ 0.93) & 61.84 ($\pm$ 25.84) \\
& 2 & $\kappa=10$ & 25.19 ($\pm$ 1.10) & 5.16 ($\pm$ 3.58) \\
& 3 & $\kappa=30$ & 24.88 ($\pm$ 1.62) & 2.95 ($\pm$ 2.64) \\
& 4 & $\kappa=60$ & 24.71 ($\pm$ 1.08) & 1.28 ($\pm$ 1.49) \\
& 5 & $\kappa=120$ & 24.13 ($\pm$ 1.55) & 0.96 ($\pm$ 1.35) \\
\midrule
\multirow{1}{*}{{N-P3O}} & 1 & $\kappa=1$ & 24.13 ($\pm$ 1.14) & 0.49 ($\pm$ 0.88) \\
\midrule
\multirow{5}{*}{{PPO-Lagrangian}} & 1 & $\lambda_{init}=0, \alpha_{\lambda}=0.001$ & 1.69 ($\pm$ 2.35) & 0.02 ($\pm$ 0.17) \\
& 2 & $\lambda_{init}=-0.5, \alpha_{\lambda}=0.0$ & 1.81 ($\pm$ 2.69) & 0.06 ($\pm$ 0.56) \\
& 3 & $\lambda_{init}=-1.5, \alpha_{\lambda}=0.0$ & 25.05 ($\pm$ 0.90) & 4.42 ($\pm$ 2.91) \\
& 4 & $\lambda_{init}=-1.4, \alpha_{\lambda}=0.001$ & 23.70 ($\pm$ 1.45) & 1.08 ($\pm$ 1.40) \\
& 5 & $\lambda_{init}=-1.3, \alpha_{\lambda}=0.001$ & 23.68 ($\pm$ 1.87) & 0.99 ($\pm$ 1.31) \\
\midrule
\multirow{5}{*}{{N-IPO}} & 1 & $\epsilon=0.3, k=20, \lambda_{rec}=1$ & 24.97 ($\pm$ 1.35) & 2.64 ($\pm$ 2.43) \\
& 2 & $\epsilon=0.2, k=20, \lambda_{rec}=1$ & 24.64 ($\pm$ 1.66) & 2.95 ($\pm$ 2.62) \\
& 3 & $\epsilon=0.1, k=20, \lambda_{rec}=1$ & 24.67 ($\pm$ 0.84) & 1.33 ($\pm$ 1.69) \\
& 4 & $\epsilon=0.05, k=20, \lambda_{rec}=1$ & 22.19 ($\pm$ 2.25) & 1.69 ($\pm$ 1.77) \\
& 5 & $\epsilon=0.025, k=40, \lambda_{rec}=1$ & 22.52 ($\pm$ 1.65) & 1.24 ($\pm$ 1.50) \\
\midrule
\multirow{5}{*}{{CRPO}} & 1 & $\epsilon=0.2$ & 24.97 ($\pm$ 1.19) & 7.36 ($\pm$ 4.88) \\
& 2 & $\epsilon=0.1$ & 24.75 ($\pm$ 1.26) & 5.13 ($\pm$ 3.28) \\
& 3 & $\epsilon=0.05$ & 24.25 ($\pm$ 1.72) & 2.65 ($\pm$ 2.18) \\
& 4 & $\epsilon=0.025$ & 23.28 ($\pm$ 1.58) & 1.62 ($\pm$ 1.68) \\
& 5 & $\epsilon=0.01$ & 22.28 ($\pm$ 1.70) & 0.96 ($\pm$ 1.22) \\
\midrule
\multirow{5}{*}{{FOCOPS}} & 1 & $\nu=1, \alpha_{\nu}=0, \lambda_{}=0.5$ & 4.59 ($\pm$ 3.76) & 0.10 ($\pm$ 0.83) \\
& 2 & $\nu=0.5, \alpha_{\nu}=0, \lambda_{}=0.5$ & 3.02 ($\pm$ 3.18) & 0.02 ($\pm$ 0.17) \\
& 3 & $\nu=0.25, \alpha_{\nu}=0, \lambda_{}=0.5$ & 2.63 ($\pm$ 3.10) & 0.03 ($\pm$ 0.21) \\
& 4 & $\nu=0.1, \alpha_{\nu}=0, \lambda_{}=0.5$ & 22.65 ($\pm$ 3.02) & 15.82 ($\pm$ 11.67) \\
& 5 & $\nu=0.1, \nu_{max}=0.2, \alpha_{\nu}=0.005, \lambda_{}=0.5$ & 2.54 ($\pm$ 3.13) & 0.11 ($\pm$ 1.45) \\
\bottomrule
\end{tabular}
\caption{Mean performance metrics and parameter values of CMDP algorithms with different parameters.}
\label{tab_comp_cmdp_algos}
\end{table*}

\section{EXPERIMENTAL RESULTS}

We present two experimentals:
\begin{enumerate}

    \item \textbf{Comparison of first-order \ac{CMDP} algorithms}:
    We select the most suitable algorithm for our purposes (N-P3O) based on a comparative study of different first-order CMDP algorithms.
        
    \item \textbf{Sim-to-real transfer with tight constraints}:  
    We validate the CMDP framework by training a perceptive locomotion policy for the robot depicted in Fig.~\ref{fig1} while enforcing tight physical constraints. 
    We compare it to a standard PPO-trained policy to assess if constrained RL offers improved constraint consistency with qualitatively similar performance.
        
\end{enumerate}



\subsection{Comparing different CMDP Algorithms}
\label{Comp_CMDP}

\subsubsection{Experimental Setup}
We consider an example problem of legged locomotion on flat terrain with constrained joint velocities. We use the ANYmal C robot and constrain the joint velocities to be below \unit[6.0]{rad/s}.

We implement all algorithms with normalized advantages, but include P3O in our comparison to depict the benefits of normalization.
%
As we aim to obtain zero constraint violations, we used P3O, N-P3O, PPO-Lagrangian and FOCOPS with a threshold ($\epsilon$) of zero. 
Hereby, the cost return cannot drop below zero since the cost function is non-negative. For N-IPO and CRPO, we treat the threshold as a hyperparameter.\footnote{
CRPO only applies reward improvement steps if the cost returns are below $\epsilon$, and the logarithm barrier penalty term in N-IPO also needs constraint satisfaction to be well-defined. }
It should be noted that a zero threshold leads to a continuous increase in the penalty parameters of PPO-Lagrangian and FOCOPS with positive learning rate.

\subsubsection{Results}
Fig.~\ref{fig:cost_rwd} and Table \ref{tab_selected_cmdp_algos} show the cost and reward over the learning iterations and the final performance of the best runs.
We include PPO without considering the constraint as a baseline. 

Three algorithms could achieve high reward and less than a single constraint violation on average: P3O, N-P3O and PPO-Lagrangian.
The N-P3O achieved the lowest constraint violation. 
Its superiority over P3O can be attributed to the balance between the reward and cost advantages due to normalization. 
With similar modification, N-IPO demonstrated the highest reward, albeit with a higher violation rate compared to P3O.
The constraint violation is unavoidable due to the non-negative $\epsilon$ by design, but potential improvements could be explored by using different cost functions and advanced scheduling techniques, as proposed by Kim et al.~\cite{kim2023not}.

\subsubsection{\textbf{Our choice}}
For our real-world experiment, we decided to use N-P3O.
Among the compared algorithms, N-P3O required the fewest parameters to adjust in our setup (with $\epsilon$ fixed at zero) and achieved low constraint violation.
Although N-IPO resulted in the highest reward and comparable constraint violation, its sensitivity to the threshold parameter made it less suitable. 
For further details on our parameter adjustments and results, please refer to Table~\ref{tab_comp_cmdp_algos} and implementation details in appendix.

\subsection{Robot Experiments }\label{robot_exp}

We evaluate a perceptive locomotion policy trained using N-P3O for our wheeled-legged robot.
We compare it with the PPO baseline trained with the constraint reward (see Table~\ref{tab:rewards}).

\subsubsection{Experimental Setup}
The policies are trained to follow velocity commands over rough terrain. 
The policy observes the terrain scan around the robot as shown by Fig.~\ref{fig1} and outputs joint position and wheel speed commands.
We used the rough terrain environment by Rudin et al.~\cite{rudin2022learning}.
The velocity commands are sampled uniformly within the ranges of [-2.0, 2.0] \unit[]{m/s} in the $x$-direction, [-1.0, 1.0] \unit[]{m/s} in the $y$-direction, and a yaw rate from [-1.5, 1.5] \unit[]{rad/s}.

To evaluate the effectiveness of the constrained RL approach, we enforce tight constraints for the leg actuators.
We use joint speed limit of \unit[6.0]{rad/s}, which is significantly lower than the robot's actual physical limit of $\sim$\unit[8.0]{rad/s}. Joint torque is limited to \unit[75]{Nm} for leg joints. The physical limit is $\sim$\unit[100]{Nm}.

We also applied other constraints mentioned in section~\ref{sec:cmdp}. We used two cost critic networks - one for command smoothness constraint and the other one for sum of other costs.

\subsubsection{Results}
In Fig.~\ref{fig_robot} we show the results from different policies in two scenarios. Both policies violated joint velocity and torque constraints at varying rates in our experiments, while other constraints remained satisfied.

Firstly, we evaluate the policies' behavior when encountered by discrete obstacles (Fig.~\ref{fig_robot}A).
A notable qualitative difference in behavior is observed: the N-P3O policy slows down before stepping down to reduce impact, while normal PPO policy gains speed (See Fig.~\ref{fig_robot}A-1(a) and A-2(a)).
This significantly impacts the rate of constraint violation.

The N-P3O policy shows two short peaks in the joint velocity that violates constraints, but the joint torque remains within the constrained range (Fig.~\ref{fig_robot}A(b)).
On the other hand, the PPO policy exhibits a significantly higher violation rate when stepping up (the front wheel collision) and when stepping down (front legs drop). 
The N-P3O policy actively modulates its leg motions and speed in response to discrete events.

Secondly, we evaluate the constraint violation when the robot is stepping at it's maximum speed to the $y$-direction (Fig.~\ref{fig_robot}B).
We commanded \unit[1.0]{m/s}, which is the maximum speed the policy is trained for.
Note that for ANYmal C robot, this is higher than the nominal operating range ($\sim$\unit[0.75] {m/s} by~\cite{lee2020learning,miki2022learning}).

As shown in Fig.~\ref{fig_robot}B-1, the N-P3O policy shows longer strides and slower gait frequency, resulting in less joint velocity constraint violation (Fig.~\ref{fig_robot}B-2). 
Additionally, the N-P3O exhibited lower tracking error. The tracking errors are $\unit[0.276 \ (\pm 0.077)]{m/s}$ and  $\unit[0.296 \ (\pm 0.091)]{m/s}$ for N-P3O and PPO, respectively. Both policies could not achieve \unit[1.0]{m/s} due to the hardware limitation.



\section{CONCLUSION \& DISCUSSION}
Our study presents a CMDP formulation for the perceptive locomotion of quadrupedal robots.
Through a comparative study of five first-order CMDP algorithms, we identified N-P3O, a normalized version of P3O, as the most effective for our task.
The additional advantage normalization step further enhanced both the stability and performance of the algorithm. 

Real-world experiments on a wheeled-legged quadrupedal robot provide strong evidence for the effectiveness of the constrained RL approach. 
Utilizing the N-P3O algorithm, our policies were able to achieve performance metrics on par with conventional PPO algorithm used by state-of-the-arts, but with fewer constraint violations. 
A distinct advantage we observed was the decoupling of reward and constraint functions, which simplified the tuning processes and led to a better performance in terms of constraint violation.

In conclusion, Constrained RL emerges as a promising tool for robotic applications, particularly in sim-to-real transfer scenarios.
While our focus was on legged locomotion, the methodology is broadly applicable.

\subsection{Practical Benefits}
From a hands-on perspective, the constrained RL algorithms showed clear advantages.
The PPO approach necessitated complex adjustments to the scaling coefficients of penalty terms (see Table~\ref{tab:rewards}).
The impact of each coefficient is non-intuitive, often demanding numerous trial-and-errors.
On the other hand, with separate cost critics, this effort is removed by design.
We can control the influence of the cost objective using a single parameter $\kappa$.
Such a streamlined approach accelerates the overall development of learned controllers. 
While having additional cost critics adds a computational overhead in comparison to PPO (\unit[0.07]{s} more), this is negligible compared to the simulation time ($\sim \unit[0.74]{s}$).

\begin{table*}
    \centering
    \begin{tabular}{ccccc}
    \hline
    Type & Definition & Reward & $\#_{\text{violations}}$ / episode & Deviation \\
    \hline
    Indicator ($\mathds{1}$) & $1$ if any constraint is violated, $0$ otherwise & 24.15 ($\pm$ 1.48) & \cellcolor{green!25}0.53 ($\pm$ 1.11) & 1.02 ($\pm$ 1.27) \\
    Number of Joints & $\sum_{i \in \{joints\}} \mathds{1}_{(\dot{q}_i > \dot{q}_{\text{max}})}$ & 24.12 ($\pm$ 1.23) & \cellcolor{green!25}0.54 ($\pm$ 0.91) & 1.13 ($\pm$ 1.52) \\
    ReLU & $\sum_{i \in \{joints\}} \max(0, \dot{q}_i - \dot{q}_{\text{max}})$ & 24.23 ($\pm$ 1.60) & 0.83 ($\pm$ 1.30) & 0.99 ($\pm$ 1.39) \\
    ReLU$^2$ & $\sum_{i \in \{joints\}} \max(0, \dot{q}_i - \dot{q}_{\text{max}})^2$ & 24.71 ($\pm$ 1.09) & 2.77 ($\pm$ 2.99) & \cellcolor{green!25}0.64 ($\pm$ 1.04) \\
    \hline
    \end{tabular}
    \caption{Mean performance metrics with different cost functions.}
    \label{tab:cost_functions}
\end{table*}

\subsection{Future work}
Future works will include different applications such as autonomous navigation or manipulation. Additionally, we only experimented with simple and constant constraints. 
More complex systems, such as joints with variable gear ratios, may introduce state-dependent constraints. Identifying complex constraints from an unknown or under-modeled systems remains an open question.
Current approaches also face limitations in enforcing hard constraints. Constraint violation is inevitable due to the exploration during training.
This issue is particularly relevant for safety-sensitive applications, necessitating the development of methods for stricter constraint satisfaction~\cite{yang2022safe, chen2021safety}. 

\section*{APPENDIX}
\setcounter{subsection}{0}
 

Here we provide additional experiments and technical details.


\subsection{Effect of Different Cost Functions}

\label{Effect_cost_fcts}
The cost function is an important design choice when formulating a CMDP. We evaluate the effect of the cost functions in Table~\ref{tab:cost_functions}, again on the example problem of quadrupedal locomotion on flat terrain with constrained joint velocities. The policies were obtained with N-P3O and $\kappa = 1$.
There are notable differences in the constraint violations. The indicator function leads to the fewest violations, closely followed by the number-of-joints cost function. The squared-ReLU cost function violates more often, but leads to smaller deviation from the limit. 

%
%
%


\subsection{Non-negative Cost Critics}
\begin{figure}[h]
    \centering
   \includegraphics[width=\linewidth]{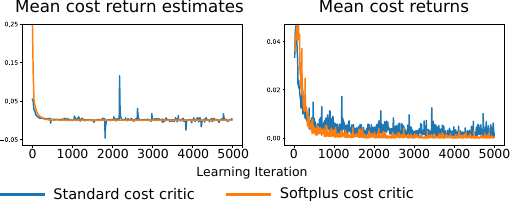}       
    \caption{Comparison of standard and softplus cost critic.}
    \label{fig:cost_returns}
\end{figure}

In cases of near-perfect constraint satisfaction, a plain cost critic has trouble learning the cost value function, often outputting negative values.
To address this, we appended a Softplus output layer to the cost critic.
Fig.~\ref{fig:cost_returns} display the mean of the sampled cost returns and the estimated cost returns.
The use of the non-negative function leads to a lower variance in cost returns.
These improvements are shown in Table~\ref{tab_compcritics}.

\begin{table}[h]
	\centering
		\begin{tabular}{cccc}
			\toprule
			Output layer & Reward & $\#_{\text{violations}}$ per episode \\
			\midrule
			Linear & 73.12 & 0.28 \\
			Softplus & 72.83 & \cellcolor{green!25}0.05 \\
			\bottomrule
		\end{tabular}%

	\caption{Mean performance metrics of the policies for torque-constrained locomotion.}
	\label{tab_compcritics}
\end{table}

\subsection{Learning Time}
Table \ref{tab_time_1} shows the time requirements per iteration of PPO and N-P3O for the velocity-constrained locomotion problem.
Training of a separate cost critic leads to an increase in learning time.
In our setting, this is negligible as the total training time is dominated by the simulation time.

\begin{table}[h]
	\centering
		\begin{tabular}{ccc}
			\toprule
			   & PPO & N-P3O\\
			\midrule
			Policy update & 0.118 ($\pm$0.006) s & 0.185 ($\pm$ 0.009) s \\
			Data collection & 0.696 ($\pm$0.014) s & 0.735 ($\pm$0.011) s\\
			\bottomrule
		\end{tabular}%
	\caption{Time requirements of PPO and N-P3O trainings.}
	\label{tab_time_1}
\end{table}



\subsection{Training Details}
The definition of observation and domain randomization are the same as Rudin et al.~\cite{rudin2022learning}.
\subsubsection{Architecture}
The models are depicted in Table~\ref{tab:arch}. The proprioceptive observation includes target velocity, base velocity, joint position, joint velocity, and gravity vector. 
\begin{table}
\centering
 \begin{tabular}{|c|c|c|c|c|}
\hline
Layer & \multicolumn{2}{|c|}{Policy \& Reward critic} &  \multicolumn{2}{|c|}{Cost critic} \\ \hline
input & proprio.* & height scans & proprio.* & height scans \\ \hline
1 & id & ELU(128) & id & ELU(128)\\ \hline
2 & id & ELU(64) & id & ELU(64)\\ \hline
3 & \multicolumn{2}{|c|}{concatenate } & \multicolumn{2}{|c|}{concatenate }\\ \hline
4 & \multicolumn{2}{|c|}{ ELU(256) } & \multicolumn{2}{|c|}{ ELU(128) }\\ \hline
5 & \multicolumn{2}{|c|}{ELU(64)  } & \multicolumn{2}{|c|}{ Output }\\ \hline
6 & \multicolumn{2}{|c|}{Output} & \multicolumn{2}{|c|}{ - }\\ \hline
 \end{tabular}
 \caption{Neural network architectures. ELU denotes fully connected layer with ELU activation function. ($*$: proprioceptive observations)}\label{tab:arch}
\end{table}

\subsubsection{Scheduling Constraint Minimization}
When constraints are enforced, we noticed premature convergence of the policy training. 
To promote exploration, we set $\kappa$ to a low value at the beginning of the training and exponentially increased the value: $\kappa_i = \min(0.2, 0.1 \cdot (1.0004)^i)$ for $i$-th iteration.

\subsubsection{Decaying Entropy Coefficient}
We introduced a decaying entropy regularization loss in the objective function, as suggested in previous work \cite{williams1992simple}, \cite{mnih2016asynchronous}.
This improved the smoothness of the policy on convergence.

\subsection{Algorithm Implementation Details}
\subsubsection{PPO-Lagrangian}
In our implementation, we use the ADAM optimizer for update in Eq.~\ref{eq:lambda_update_ppolagr} and apply the Softplus function to ensure non-negativity of $\lambda$ after updates.

\subsubsection{N-IPO}
The logarithm barrier penalty cannot be applied if the constraint is already violated. 
We added a recovery strategy to achieve constraint satisfaction again:
\begin{align*}
    L^{\text{N-IPO}}(\theta') &= L_R^{\mathrm{CLIP, N}}(\theta') 
    + \sum_{i: J_{C_i}(\theta') \leq \epsilon_i} \phi(L_{C_i}^{\text{VIOL,N}}(\theta'))  \\
    &+ \lambda_{\text{rec}} \cdot \sum_{i: J_{C_i}(\theta') > \epsilon_i} L_{C_i}^{\mathrm{CLIP,N}}(\theta')
\end{align*}
 with the additional recovery term.

\subsubsection{CRPO}
We utilize sampled data within multiple epochs and iterate over minibatches, leading to several updates of the policy within each learning iteration. In our implementation of CRPO, we utilize the constraint reformulation (Eq. \ref{eq_reformConstr}) to estimate the constraint violation after every policy update, instead of switching between policy improvement and constraint minimization after each complete iteration.







\bibliographystyle{bibliography/IEEEtran}

\end{document}